# Vertebral body segmentation with *GrowCut*: Initial experience, workflow and practical application

Jan Egger[1,2,3,4], Christopher Nimsky[3] and Xiaojun Chen[5]

## Abstract

**Objectives:** Spinal diseases are very common; for example, the risk of osteoporotic fracture is 40% for White women and 13% for White men in the United States during their lifetime. Hence, the total number of surgical spinal treatments is on the rise with the aging population, and accurate diagnosis is of great importance to avoid complications and a reappearance of the symptoms. Imaging and analysis of a vertebral column is an exhausting task that can lead to wrong interpretations. The overall goal of this contribution is to study a cellular automata-based approach for the segmentation of vertebral bodies between the compacta and surrounding structures yielding to time savings and reducing interpretation errors.
**Methods:** To obtain the ground truth, T2-weighted magnetic resonance imaging acquisitions of the spine were segmented in a slice-by-slice procedure by several neurosurgeons. Subsequently, the same vertebral bodies have been segmented by a physician using the cellular automata approach *GrowCut*.
**Results:** Manual and *GrowCut* segmentations have been evaluated against each other via the Dice Score and the Hausdorff distance resulting in 82.99% ± 5.03% and 18.91 ± 7.2 voxel, respectively. Moreover, the times have been determined during the slice-by-slice and the *GrowCut* course of actions, indicating a significantly reduced segmentation time (5.77 ± 0.73 min) of the algorithmic approach.
**Conclusion:** In this contribution, we used the *GrowCut* segmentation algorithm publicly available in three-dimensional Slicer for three-dimensional segmentation of vertebral bodies. To the best of our knowledge, this is the first time that the *GrowCut* method has been studied for the usage of vertebral body segmentation. In brief, we found that the *GrowCut* segmentation times were consistently less than the manual segmentation times. Hence, *GrowCut* provides an alternative to a manual slice-by-slice segmentation process.

## Introduction

The human back is an intricate structure consisting of bones, muscles and other tissues. It is the posterior part of the body's trunk and it ranges from the neck to the pelvis.[1] A large part of the bones are the vertebral bodies that lie anterior to the spinal cord. Anyway, osteoporosis, which is a decrease in bone density and mass, is a major public health concern, and recent clinical and epidemiologic trials on osteoporosis proved the need for a precise recognition and diagnosis of vertebral fractures.[2] In total, 30 million American women and 14 million American men are affected by osteoporosis or osteopenia, and the overall lifetime risk for an osteoporotic fracture is 40% in White women and 13% in White men in

[1]Institute of Computer Graphics and Vision, Graz University of Technology (TUG), Graz, Austria
[2]BioTechMed-Graz, Graz, Austria
[3]Department of Neurosurgery, University Hospital Marburg, Marburg, Germany
[4]Computer Algorithms for Medicine (Cafe) Laboratory, Graz, Austria.
[5]School of Mechanical Engineering, Shanghai Jiao Tong University, Shanghai, China

**Corresponding authors:**
Jan Egger, Institute of Computer Graphics and Vision, Graz University of Technology (TUG), Inffeldgasse 16c/2, 8010 Graz, Austria.
Email: egger@tugraz.at

Xiaojun Chen, School of Mechanical Engineering, Shanghai Jiao Tong University, Dongchuan Road 800, Minhang District, 200240 Shanghai, China.
Email: xiaojunchen@163.com; xiaojunchen@sjtu.edu.cn

the United States.[3,4] In general, a vertebral fracture affects daily living activities, including getting up from a chair, walking, taking stairs, bathing, dressing and cooking.[5–7] The increased number of spinal surgical procedures among older patients results from an increasing incidence for vertebral bone diseases, which is often responsible for a limited mobility and overall quality of life. However, when a decision for an adequate treatment is made, neuro-imaging plays an important role for the estimation of the treatment dimension, like surgery.[8] In this decision process, an accurate and objective analysis of the vertebral deformities is very important for the diagnosis and a following therapy.[9]

Here, a computer-assisted diagnosis (CAD) system has the aim to reduce interpretation errors produced by the exhausting tasks of image screening and radiologic diagnosis. Furthermore, a CAD system can assist in the characterization and the quantification of abnormalities.[10] However, an (automatic) segmentation of vertebral bodies in magnetic resonance imaging (MRI) acquisitions is a demanding task, among other things because of the diversification in soft tissue.[11] In summary, there are three goals for this publication: the first goal is to introduce and demonstrate the segmentation of vertebral bodies with *GrowCut*, the second goal is to apply the *GrowCut* segmentation to a set of vertebral bodies from T2-weighted MRI acquisitions from the clinical routine and the third goal is to show that a *GrowCut*-based segmentation is faster than a manually slice-by-slice segmentation but at the same time can achieve a similar segmentation accuracy. Note, that the vertebral bodies in this contribution have been segmented between the compacta and surrounding structures.

In general, manual segmentations are prone to errors, because of inter- and intra-subject variabilities and the subjective judgment that is employed. Thus, the use of computer vision methods is an attractive alternative by means of providing an automation for vertebrae segmentation.[12] However, up-to-date manual segmentations in medical image analysis are still considered as ground truth, because they incorporate expert knowledge gained over several years in this area. In addition, humans can still handle exceptional changes in the images. Examples are missing slices, different image appearances because of a varying MRI sequences or a lateral instead of the standard supine position during image acquisition. An automatic segmentation of a structure in a "standard" image acquisition is already challenging and still under active research, but such exceptions would definitely overstrain most available algorithms. Hence, segmentation algorithms are in general evaluated and compared with the ground truth of manual segmentations. Several approaches have been proposed in the literature for performing (medical) image segmentation, like deformable models,[13] machine learning techniques[14] or graph-based approaches.[15] For the segmentation of spinal columns of MRI datasets in two-dimensional (2D), different graph-based algorithms have been applied, like the disk tracker algorithm that used single cross section of spinal column to achieve the segmentation,[16] and Klinder et al.[17] proposed an automated model-based vertebra detection, identification and segmentation approach.[18] Huang et al.[19] presented a fully automatic vertebra detection and segmentation system. In brief, the system consists of three stages: (1) an AdaBoost-based vertebra detection, (2) the detection refinement via robust curve fitting and (3) the vertebra segmentation by an iterative normalized cut algorithm. Michopoulou et al.[20] present an atlas-based segmentation algorithm for intervertebral disks and reports dice similarity indexes between 87% and 92%. Carballido-Gamio et al.[11,21] used normalized cuts (N-cuts) with the Nyström approximation method as segmentation technique and applied it to vertebral bodies from sagittal T1-weighted magnetic resonance (MR) images of the spine. Peng et al.[22] proposed a method using an intensity profile on a polynomial function for automated spinal detection and segmentation. Egger et al.[8] published a study on 2D segmentation of vertebrae using a rectangle-based algorithm. They also compared their approach with a *GrowCut*-based segmentation, however, only in 2D. A fully automated three-dimensional (3D) segmentation method for MR acquisition of the human spine, that uses statistical shape analysis and template matching of gray-level intensity profiles, has been introduced by Neubert et al.[23] Ghebreab and Smeulders[24] described a deformable integral spine model for segmentation. Thereby, they encode the statistics into a necklace model, on which landmarks are differentiated on their free dimensions. Stern et al.[25] segment vertebral bodies in MR and computed tomography (CT) images with deterministic models in 3D, which are initialized with a single point inside the vertebral body. Weese et al.[26] combine active shape models and elastically deformable models, by embedding the active shape model into the elastically deformable surface model. Hoad and Martel[27] developed a method for computer-assisted surgery of the spine that separates bone from soft tissue in MR images. Their segmentation approach combines thresholded region growing with morphological filtering and masking using set shapes. Yao et al.[28] describe an algorithm for automated spinal column extraction and partitioning. They start with thresholding to obtain an initial spine segmentation, followed by a hybrid method based on the watershed algorithm and directed graph search for extraction of the spinal canal. In addition, a four-part vertebra model consisting of the vertebral body, the spinous process and the left/right transverse processes is fitted for segmentation of the vertebral region and separated it from adjacent ribs and other structures.

In this contribution, an interactive version of the cellular automata algorithm called *GrowCut* was applied to the segmentation of vertebral bodies in 3D (preliminary results have been presented at the *spine* congress of the DGNC in Frankfurt, Germany[29] and as SPIE poster[30]). In a nutshell, it was discovered that a semi-automatic segmentation with *GrowCut* can achieve a similar accuracy as pure manual slice-by-slice segmentations while contemporaneously reducing the segmentation time. To perform a statistic evaluation of the *GrowCut*-based segmentation results, vertebrae images in MRI acquisition from the clinical routine have been used. Thereby, the vertebral bodies were manually outlined on a slice-by-slice basis by several physicians and a pure manual segmentation of a single vertebra (which still represents the state-of-the-art in clinics), took in average of over 10 min (10.75 ± 6.65). The ground truth for the evaluation was generated by physicians (neurological surgeons) who have all several years of practical experience in spine treatment, especially in spine surgery. As result, the direct statistical comparison of the semi-automatic *GrowCut*-based segmentations to the pure manual slice-by-slice neurological surgeons segmentations, yielded to an average Dice Similarity Coefficient (DSC)[31,32] of 82.99% ± 5.03% and in addition to an average Hausdorff distance of 18.91 ± 7.2 voxel.

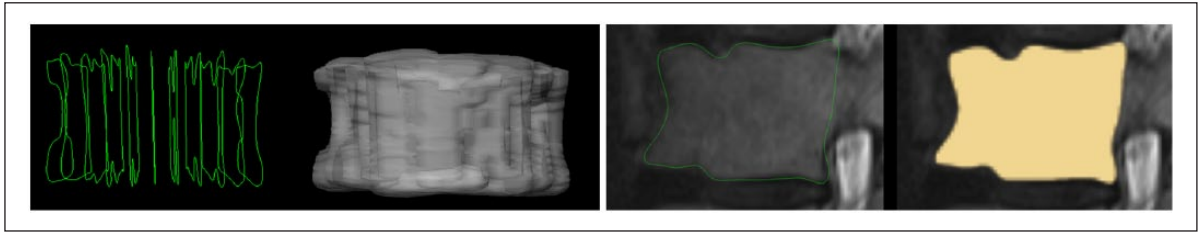

**Figure 1.** Voxelizations in 2D and 3D: the leftmost image shows several manual contours (green) of a vertebral body and the corresponding voxelized vertebra mask (gray) in the next image to the right side. The third image from the left side shows one single manual contour of a vertebral body (green) in a sagittal slice, and the rightmost image presents the corresponding voxelized mask (yellow).

The publication is organized at follows: The next section presents the "Materials and methods" used in this study, then the "Results" section is presented and finally, the "Conclusion and discussion" section discusses the contribution and outlines areas of future work.

## Materials and methods

### Data

In total, 13 vertebral bodies from different subjects have been segmented in three diagnostic T2-weighted MRI scans of the vertebral column for this study. All datasets have been acquired on a MAGNETOM Sonata scanner (1.5 Tesla MRI) from Siemens with 4-mm slice thickness. However, for a consistent comparison and evaluation, the datasets have been reformatted afterward to isotropic resolutions: twice to $0.63 \times 0.63 \times 0.63$ mm³ and once to $0.73 \times 0.73 \times 0.73$ mm³, resulting in sizes of $512 \times 512 \times 113$, $512 \times 512 \times 113$ and $512 \times 512 \times 70$ voxels. This is a retrospective study with anonymized image data, which does not require an ethics approval. Thus, no written consent was needed by the patients. In addition, the datasets are freely available for download and have previously been used in the contributions Egger et al.:[8] and Zukic et al.:[33] http://www.cg.informatik.uni-siegen.de/de/spine-segmentation-and-analysis and https://www.researchgate.net/publication/287214481_Spine_Datasets (last accessed on October 2017).

### Manual outlining

Each vertebral body has been manually outlined on a slice-by-slice basis by neurosurgeons of University Hospital of Marburg (UKGM) in Germany (Chairman: Professor Dr C.N.). The 3D manual slice-by-slice segmentations were performed in the sagittal direction (with corrections in axial and coronal directions were necessary), and all neurosurgeons had several years of experience in the treatment of vertebral diseases. However, if the border of the vertebra was very similar between consecutive (sagittal) slices, the contouring software allowed the user to skip the manual segmentation for these slices. Instead, the vertebral boundaries were interpolated by the contouring software in these areas. The basic contouring software, used for the manual contouring process, was established with a network under the medical prototyping platform MeVisLab (www.mevislab.de, date of access: October 2017)[34–36] running on an up-to-date laptop with Microsoft Windows. However, beside the simple contouring capabilities, the software provided no algorithmic support to avoid falsifying the segmentation outcomes.

### Software

Similar to Egger et al.[37] for pituitary adenoma,[38] for glioblastoma multiforme (GBM) and[39] for lung cancer, the algorithm used for this segmentation study is implemented open source within (3D) Slicer (http://www.slicer.org).[40,41] Briefly, Slicer is a medical image computing platform for biomedical research that can freely be used for other research studies. The ground truth for our study has been acquired via manual slice-by-slice segmentations of the vertebral bodies as described in the previous section. Voxelizations of manual segmentations in 2D and 3D are presented in Figures 1 and 2, respectively. The leftmost image of Figure 1 shows manual contours (green) and the corresponding voxelized vertebra mask (gray) is presented in the next image to the right side. The third image of Figure 1 from the left side shows a single manual contour of a vertebral body (green) in a sagittal slice and the rightmost image of Figure 1 presents the corresponding voxelized mask (yellow). Finally, in Figure 2, seven vertebral bodies that have been manually segmented are voxelized (first three images from the left) and superimposed in 3D (yellow) on a corresponding sagittal MRI slice (these kind of manual 3D segmentation masks of vertebral bodies have been used for the statistical evaluation).

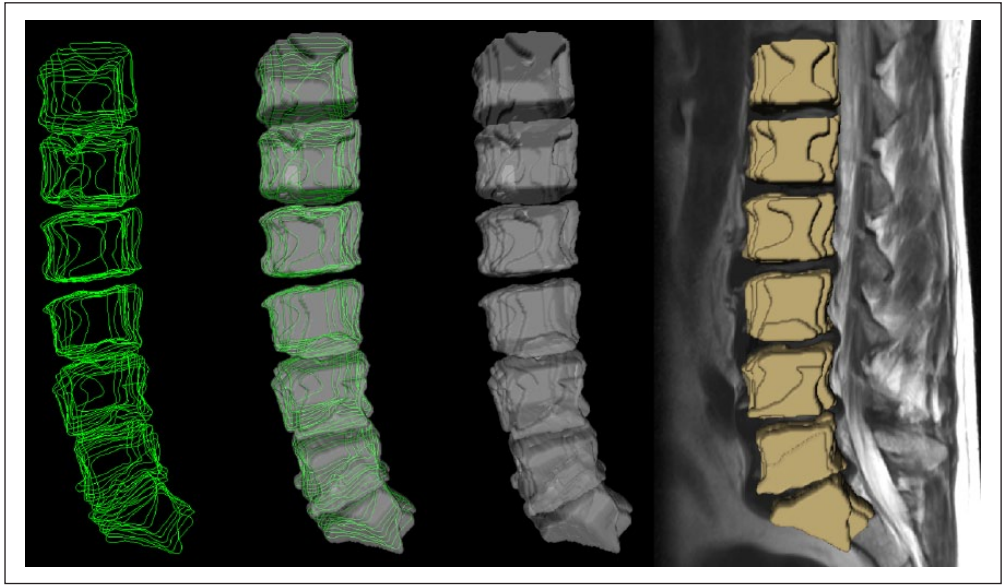

**Figure 2.** Voxelization of manual segmentations in 3D (sagittal view) used for evaluation in this study: manual outlines of seven vertebral bodies in green (left), corresponding voxelized masks in gray (middle images) and voxelized masks superimposed with a sagittal MRI slice in yellow (rightmost image).

## GrowCut

The segmentation algorithm *GrowCut* is a competitive region-growing approach that uses cellular automata as an iterative labeling procedure. As already shown in previous studies,[37–39] a *GrowCut*-based segmentation is able to achieve reliable and reasonably fast segmentations of moderately difficult 2D and 3D objects. Summarized, *GrowCut* initializes a cellular automata, where each cell is associated to an image pixel and its state is stored as a three-tuple (l, s, C), with l is the foreground/background label, s is the strength of the cell and C encodes the color/gray value information of the corresponding pixel. Furthermore, *GrowCut* uses as initialization a set of input pixels for the foreground l = 1 (in this study the vertebral body) and a set of input pixels for the background l = 0 provided by the user with s = 1. After the initialization, *GrowCut* labels all remaining pixels in the image iteratively either as foreground or as background and terminates when all pixels in the *Region of Interest* (*ROI*) have been assigned a label. The following pseudocode summarizes the *GrowCut* evolution rule:[42,43]

    // for all cells
        // copy previous states
        l'[p] = l[p], s'[p] = s[p]
        for all C4 or C8 neighbor q of the current cell
        if g(Dist(C[p], C[q]))s[q]>s[p] then
            // update cell state
            l'[p] = l[q], s'[q] = g(Dist(C[p], C[q]))s[q]

With Dist() returning the distance between two color/gray values and g() monotonous decreasing function guaranteeing convergence. A detailed video demonstrating the *GrowCut* evolution under Slicer can be found here (date of access: October 2017): https://www.youtube.com/watch?v=dfu2gugHLHs.

The video shows the step-by-step growing of the different labels (foreground/background), how they interact with each other and finally, stop when the final segmentation is achieved (In general, the intermediate region-growing steps are not shown to the user, rather the final segmentation outcome is provided directly to the user to speed up the overall segmentation process).

## GrowCut *implementation*

The current implementation of *GrowCut* in Slicer that has been used for this study consists of a graphical user interface (GUI) front-end and an algorithm back-end. The front-end enables interactions of the user with the image and therefore allows the user to paint directly on the image. The back-end on the other side computes the segmentation after the initialization phase. For a detailed implementation, we want to refer the reader at this point to https://www.slicer.org/slicerWiki/index.

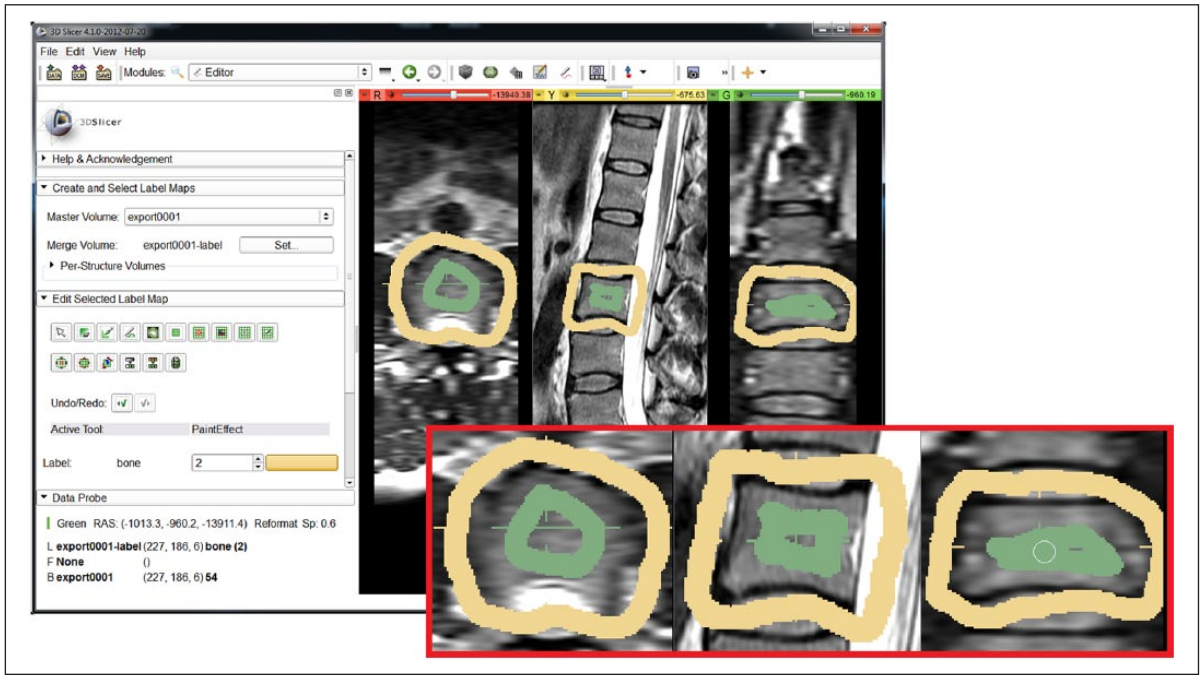

**Figure 3.** The screenshot shows a classical user initialization of *GrowCut* during this study: The *Editor* module (left) is used to mark parts of the vertebra (green) and the background (yellow) in an axial (first window), sagittal (second window) and coronal plane (third window).

php/Modules:GrowCutSegmentation-Documentation-3.6 (last accessed on October 2017).

However, the 3D Slicer implementation of the *GrowCut* approach employs some techniques to speed up the automatic segmentation process, which are introduced here:

1. The *GrowCut*-based implementation computes the segmentation only within a small area. That safes computation time and can be done, because a user is generally only interested in segmenting out a small area (in this study a single vertebral body) of the whole image. Achieved is this, by computing the segmentation area as a convex hull of the user-labeled pixels/voxels, plus an extra margin around 5%.[38]
2. Using multiple threads to execute iterations involving the image, multiple small areas of the image are also updated simultaneously within the *GrowCut*-based implementation.
3. Furthermore, the similarity distance between the pixels/voxles is pre-computed once at the beginning and then used again later.
4. Pixels/voxels that are already labeled with weights (so-called "saturated" pixels/voxels), that cannot be updated anymore, are memorized by the implementation.

### Slicer-based vertebral body segmentation

The 3D Slicer platform offers several segmentation tools for different tasks in the medical domain, like a simple region-growing approach or the Robust Statistics Segmenter (RSS).[44] However, after performing initial tests for vertebral body segmentation, the *GrowCut*-based approach, followed by optional morphological operations (like dilation, erosion and island removal), achieved the most promising segmentation results on our clinical datasets. Hence, the subsequent step-by-step workflow has been worked out for the vertebral body segmentation task, which was also used to train new users: (1) open a new patient dataset within 3D Slicer; (2) initializing *GrowCut*, by marking an area inside a vertebral body—foreground—and outside the same vertebral body—background; (3) execute *GrowCut*; (4) optional post-editing, mostly for difficult cases: apply morphological operations like dilation, erosion and island removal, after a visual inspection of the segmentation results. Finally, Figure 3 shows the overall user interface with the so-called *Editor* module on the left side and a dataset on the right side. The *Editor* module has been used for the *GrowCut* initialization and the morphological operations, and a typical initialization of a L4 vertebra on an axial, a sagittal and a coronal cross section is shown on the right side. Thereby, the foreground voxels, belonging to the vertebral body, have been marked in green and the background voxels, belonging to surrounding structures, have been marked in yellow. The hardware and software operating system that have been used for this study were an up-to-date computer (Intel Core i5-750 CPU, $4 \times 2.66$ GHz, 8 GB RAM) with Microsoft Windows (XP Professional x64 Version, Version 2003, Service Pack 2) installed.

**Table 1.** Direct comparison of manual slice-by-slice and Slicer-based *GrowCut* segmentation results for thirteen vertebral bodies via the Dice Similarity Coefficient (DSC) and the Hausdorff distance. The last column presents the time in minutes for the *GrowCut*-based segmentations.

| Vertebral body no. | Volumes of the vertebral bodies (mm³) | | Hausdorff distances (voxel) | DSCs (%) | Times (min) |
| --- | --- | --- | --- | --- | --- |
| | Manual | Slicer-based | | | |
| 1 | 49,396.2 | 42,914.7 | 32.3 | 85.17 | 7 |
| 2 | 42,196.5 | 40,256.6 | 23.35 | 88.54 | 6 |
| 3 | 42,632.3 | 45,124.8 | 22.72 | 91.6 | 6 |
| 4 | 39,419.4 | 42,260.3 | 31.49 | 85.88 | 6 |
| 5 | 29,910.6 | 27,204.8 | 12.64 | 83.87 | 5 |
| 6 | 33,908 | 35,665.3 | 18.09 | 86.45 | 5 |
| 7 | 35,492 | 46,950.7 | 15.35 | 85 | 5 |
| 8 | 39,220.7 | 54,737.8 | 10.7 | 82.62 | 6 |
| 9 | 38,653.2 | 59,216.3 | 22.47 | 78.39 | 6 |
| 10 | 39,439.4 | 55,530.6 | 10.81 | 80.73 | 5 |
| 11 | 33,107.7 | 53,288.2 | 17.12 | 74.69 | 7 |
| 12 | 30,097.7 | 43,296.8 | 12.47 | 81.4 | 6 |
| 13 | 20,888.6 | 33,535.6 | 16.27 | 74.56 | 5 |

DSC: Dice Similarity Coefficient.

## Results

Besides the segmentation time, the DSC and the Hausdorff distance have been used as comparison metrics in this study. Thereby, pure manually slice-by-slice expert segmentations (ground truth) have been evaluated against the *GrowCut*-based vertebral body segmentations. In brief, the DSC or Dice Score is a measure for spatial overlap of several segmentations of the same object and commonly applied in medical imaging studies for a quantification of the overlap degree between the segmented objects, for example, $A$ and $R$: $DSC = 2 \cdot V(A \cap R)/(V(A)+V(R))$. As result, the Dice Score can have a value ranging from *zero* to *one*, whereby a value of *zero* indicates no overlap between the segmentations and a value of *one* indicates a perfect agreement between the segmentations (as a consequence higher values indicate a better agreement). Mathematically, the DSC is defined as two times the intersection volume between the two segmentations $A$ and $R$, divided by the sum of both segmentation volumes.[45] In contrast to the Dice Score, the Hausdorff distance[46] is calculated to indicate how far away (in voxel) two segmentations $A$ and $R$ are; hence, both metrics complement one another very good. As gold standard for calculating the Dice Scores and the Hausdorff distances, manual segmentations of vertebrae boundaries were extracted. This task was performed by clinical experts (neurological surgeons) with several years of experience in spine surgery. A comparison between the manual segmentations and the *GrowCut*-based segmentations yielded to an average Dice Score of 82.99%±5.03% and a Hausdorff distance of 18.91±7.2 voxel. Thereby, the *GrowCut*-based segmentations had been performed by a physician, who had been trained in the usage of *GrowCut*. Table 1 presents the detailed results for all vertebral bodies of this study. Columns two and three indicate the volume for the vertebral bodies in mm³ for the pure manual segmentations (column two) and the *GrowCut*-based segmentations (column three). The Hausdorff distances in voxels and the DSCs between the segmentations for the single vertebral bodies are presented in the next two columns. Finally, the last column shows the times in minutes for the *GrowCut*-based segmentations. In addition, Table 2 presents the summary of the results—minimum, maximum, mean $\mu$ and standard deviation $\sigma$—for all vertebral bodies from Table 1.

Figure 4 presents a *GrowCut*-based segmented vertebral body (green) in different views (2D and 3D) for visual inspection: the upper images show the segmentation results in 2D for axial, sagittal and coronal planes. The lower left image shows a 3D view of the segmented vertebral body containing axial, sagittal and coronal planes. Furthermore, the lower image on the right side shows a 3D representation of a segmented L4 vertebral body with an additional surface smoothing under 3D Slicer. In Figure 5, a direct comparison of a manual (yellow) and a *GrowCut*-based segmentation (green) on a sagittal MRI slice is presented. Thereby, the upper left image shows the original MRI acquisition, and the upper right image presents a pure manual segmentation. The lower left image, however, shows a *GrowCut*-based segmentation result, and the lower right image shows both segmentations superimposed (manual and *GrowCut*-based).

For a direct compassion of the achieved results with other segmentation approaches, we also applied a graph-based method[47–50] and a deformable model[51] to the same clinical datasets (Table 3). The graph-based approach used a cubic-shaped template for the segmentation of the single vertebral bodies and resulted in a DSC of 81.33% and a running time

**Table 2.** Summary of segmentation results, presenting minimum, maximum, mean $\mu$ and standard deviation $\sigma$ for 13 vertebral bodies.

| | Volumes of the vertebral bodies (cm³) | | Hausdorff distances (voxel) | DSCs (%) | Times (min) |
|---|---|---|---|---|---|
| | Manual | Slicer-based | | | |
| Minimum | 20.89 | 27.2 | 10.7 | 74.56 | 5 |
| Maximum | 49.4 | 59.22 | 32.3 | 91.6 | 7 |
| $\mu \pm \sigma$ | 36.49 ± 7.15 | 44.61 ± 9.36 | 18.91 ± 7.2 | 82.99 ± 5.03 | 5.77 ± 0.73 |

DSC: Dice Similarity Coefficient.

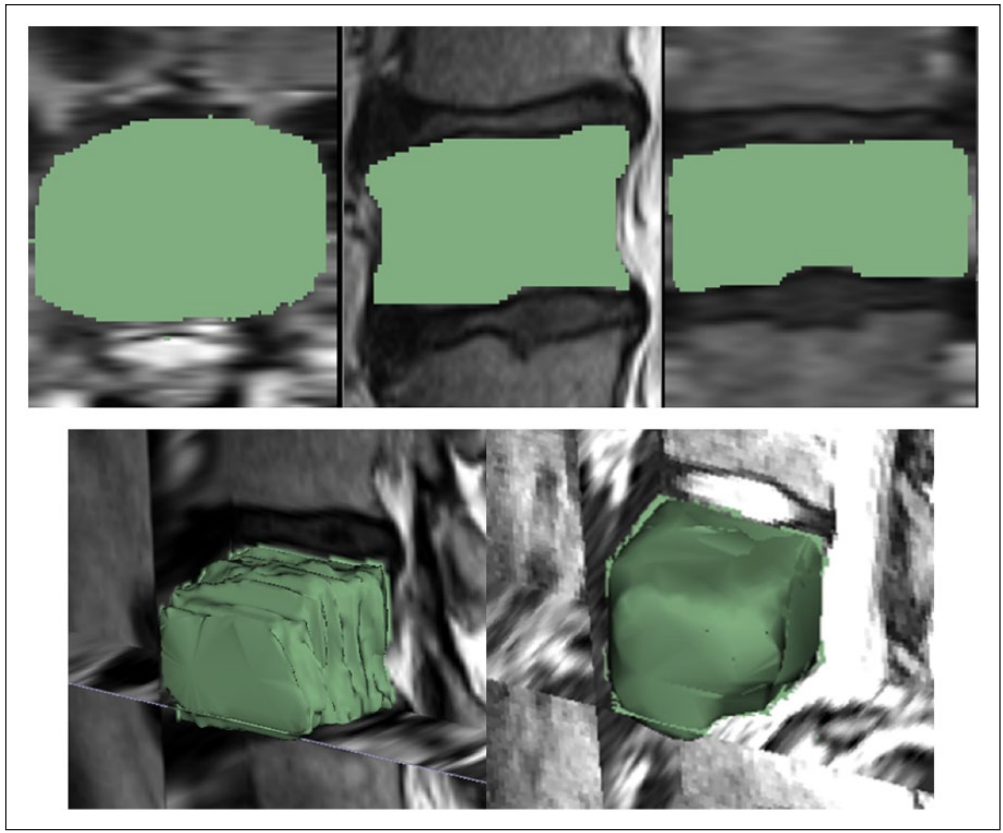

**Figure 4.** Segmentation result of a vertebral body (green) under *GrowCut*: The three upper images show the segmentation results in 2D for axial, sagittal and coronal planes, respectively. The lower left image presents a 3D view of the segmented vertebral body with axial, sagittal and coronal planes, and the lower right image shows a three-dimensional representation of a segmented L4 vertebral body with additional surface smoothing under 3D Slicer.

of less than a minute. However, several parameters had to be chosen like a smoothness term and the amount of graph nodes. The deformable model approach segmented the single vertebra using multiple-feature boundary classification and mesh inflation,[52] and it started with a simple point-in-vertebra initialization. The average Dice coefficient was 79.3%, and the average dataset processing time was about 75 s. However, once the mesh inflation for the segmentation of the single vertebra started, the user had no options to intervene and hence had to start over again when the segmentation outcome was not satisfying.

In summary, the graph-based and the deformable model approach need a point-in-vertebra initialization. Thus, the proposed deformable model approach used an automatic vertebral body detection in a first step. For the graph-based approach, the point-in-vertebra initialization was done manually. However, the manual initialization could also be replaced with the automatic vertebral body detection from the deformable model approach. At this point, we want to refer to a recent publication about a comprehensive evaluation and comparison of 3D intervertebral disk localization and segmentation methods for 3D T2 MR data.[53]

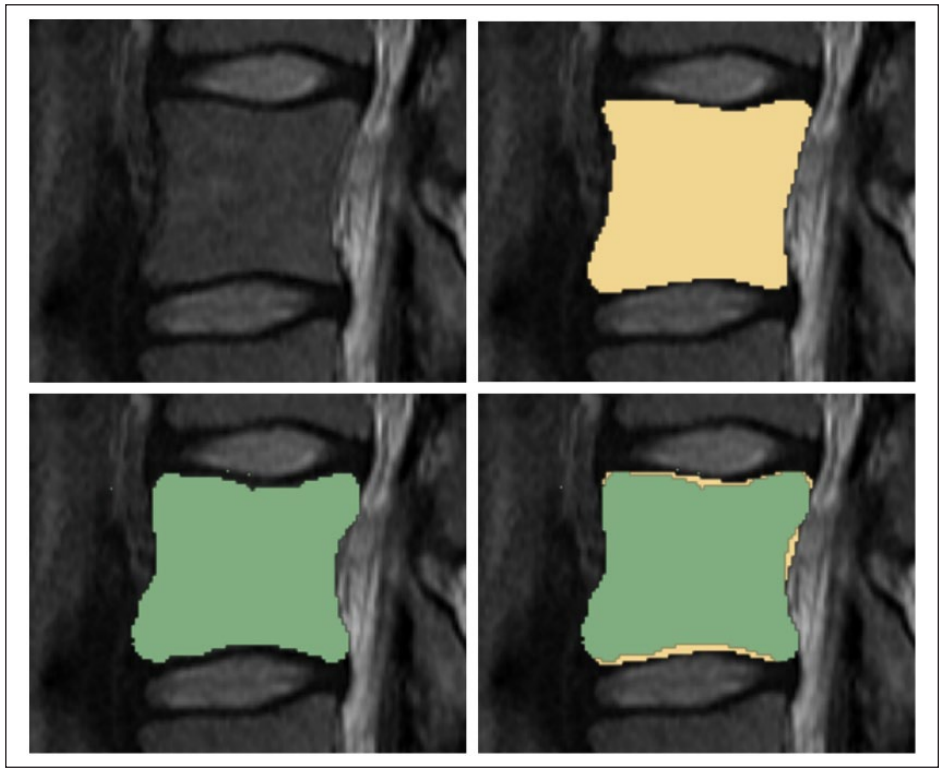

**Figure 5.** For visual inspection, this image presents a direct comparison of a manual (yellow) and the *GrowCut*-based segmentation (green) on a sagittal slice. Hence, the upper image shows on the left side the original magnetic resonance imaging acquisition and the upper image on the right side presents a corresponding manual segmentation on the same slice. Further, the lower image on the left side presents the *GrowCut-based* segmentation result and the lower image on the right side shows both segmentations superimposed (manual and *GrowCut*-based).

**Table 3.** Direct comparison of the Dice Similarity Coefficients (DSCs) between a deformable model, a graph-based and the Slicer *GrowCut* approach on the same clinical datasets.

|  | Deformable model[51] | Graph-based[47–50] | GrowCut |
| --- | --- | --- | --- |
| DSCs (%) | 79.3 | 81.33 | 82.99 |

## Conclusion and discussion

In this contribution, we used the *GrowCut* segmentation algorithm available in 3D Slicer for 3D segmentation of vertebral bodies. To the best of our knowledge, this is the first time that the *GrowCut* method of Slicer has been studied for the usage of vertebral body segmentation. In brief, we found an average segmentation time for a *GrowCut*-based segmentation of less than 6 min (5.77 ± 0.73). This is consistently less than the manual segmentation times of all 13 vertebrae we used for an evaluation. The mean Dice Similarity score was 82.99% ± 5.03% and the mean Hausdorff distance was 18.91 ± 7.2 voxel, which indicated similarity with the pure manual slice-by-slice segmentation. Summing up, for our study, a *GrowCut*-based segmentation applied to vertebral images was less time-consuming than laborious manual segmentations.

Even if the results are not perfect and there were still some manual corrections of the *GrowCut* outcome necessary for the most cases, everybody can download 3D Slicer and test it (in contrast to the most existing approaches that are only available within the research groups that developed them). In addition, no parameters have to be defined at all, and the initialization of *GrowCut* is done via an intuitive brush-based manner. Thus, this simple segmentation tool can also be used by end-users, like clinicians, which has been shown within this publication. However, a user should keep in mind that the initialization of the labels is important and requires specialist knowledge. In summary, *GrowCut* can be used for the segmentation of moderately difficult objects in 2D and 3D, because it achieves in these cases reliable and reasonably fast results. However, because of its interactive nature, it should be applied on a case-by-case basis. For processing and analyzing huge amount of datasets at once, fully automatic approaches are more suitable.

Even though we never had to reinitialize *GrowCut* on the same vertebral body, the localization of the initial labeling impacts the segmentation outcome. However, we already performed a more systematic analysis of different initializations and corresponding final segmentations in Egger et al.[8]

Despite our results, we are aware of some limitations concerning this study: first, the segmentation method assessed in this contribution is not completely new, since the already existing *GrowCut* algorithm can be used in many variations in different software applications and platforms.[54–56] Second, although the datasets used in this study were selected in random in the clinical routine, higher amount of data samples would probably have more impact on assessing the feasibility and accuracy of the used segmentation approach. Third, although ground truth generation was tried to be performed as valid as possible by clinical experts, a real image-based ground truth scheme used as comparative segmentation volume is impossible to create, since every segmentation approach has to relief on certain image-based landmarks.

There are several areas of future work, for instance, a more convenient *GrowCut* initialization. In this contribution, the initialization of the foreground and background was setup by the user in three slices (axial, sagittal and coronal), but instead, one single 3D initialization could be applied. This could be achieved by constructing two cubes around a user-defined seed point near the center of the vertebral body: a small cube that is located inside the vertebral body and marks part of the foreground, and a larger cube enclosing the vertebral body marking parts of the background. Finally, the introduced interactive segmentation could also be turned into a fully automated approach: starting with an automatic detection of the vertebral canters using a Viola–Jones-like method,[51] followed by an automatic initialization of *GrowCut*. Thereby, the automatic initialization could be performed by marking a small cubical or spherical area around the detected center (that belongs to the vertebral body) as foreground (and an automatic background initialization could be achieved via a certain distance to the center point).

### Declaration of conflicting interests

The author(s) declared no potential conflicts of interest with respect to the research, authorship and/or publication of this article.

### Funding

The author(s) disclosed receipt of the following financial support for the research, authorship, and/or publication of this article: This study was supported by the Foundation of Science and Technology Commission of Shanghai Municipality (15510722200, 16441908400), and Shanghai Jiao Tong University Foundation on Medical and Technological Joint Science Research (YG2016ZD01, YG2015MS26). Dr rer. physiol. Dr rer. nat. J.E. receives funding from BioTechMed-Graz ("Hardware accelerated intelligent medical imaging") and the 6th Call of the Initial Funding Program from the Research & Technology House (F&T-Haus) at the Graz University of Technology. Supported by TU Graz Open Access Publishing Fund.